%% file: PaperForReview.tex
\documentclass[10pt,twocolumn,letterpaper]{article}

\usepackage[pagenumbers]{cvpr}
\usepackage{graphicx}
\usepackage{amsmath}
\usepackage{amssymb}
\usepackage{booktabs}
\usepackage{multirow}

\usepackage[T1]{fontenc}
\usepackage{scalefnt}

\makeatletter
\newcommand*\bigcdot{\mathpalette\bigcdot@{.5}}
\newcommand*\bigcdot@[2]{\mathbin{\vcenter{\hbox{\scalebox{#2}{$\m@th#1\bullet$}}}}}
\makeatother

\usepackage[pagebackref,breaklinks,colorlinks]{hyperref}

\usepackage[capitalize]{cleveref}
\crefname{section}{Sec.}{Secs.}
\Crefname{section}{Section}{Sections}
\Crefname{table}{Table}{Tables}
\crefname{table}{Tab.}{Tabs.}

\def\confName{CVPR}
\def\confYear{2022}

\begin{document}

\title{Unified Multimodal Pre-training and Prompt-based Tuning \\for Vision-Language Understanding and Generation}

\author{Tianyi Liu$^{1}$, Zuxuan Wu$^{1}$, Wenhan Xiong$^{2}$,
Jingjing Chen$^{1}$, Yu-Gang Jiang$^{1}$ \\
$^1$Shanghai Key Lab of Intelligent Information Processing,\\
School of Computer Science, Fudan Univeristy\\
$^2$University of California, Santa Barbara
}
\maketitle

\begin{abstract}
Most existing vision-language pre-training methods focus on understanding tasks and use BERT-like objectives (masked language modeling and image-text matching) during pretraining. Although they perform well in many understanding downstream tasks, e.g., visual question answering, image-text retrieval and visual entailment, they do not possess the ability to generate. To tackle this problem, we propose Unified multimodal pre-training for both Vision-Language understanding and generation (UniVL). The proposed UniVL is capable of handling both understanding tasks and generative tasks. We augment existing pretraining paradigms that only use random masks with causal masks, i.e., triangular masks that mask out future tokens, such that the pre-trained models can have autoregressive generation abilities by design. We formulate several previous understanding tasks as a text generation task and propose to use prompt-based method for fine-tuning on different downstream tasks. Our experiments show that there is a trade-off between understanding tasks and generation tasks while using the same model, and a feasible way to improve both tasks is to use more data. Our UniVL framework attains comparable performance to recent vision-language pre-training methods on both understanding tasks and generation tasks. Moreover, we demostrate that prompt-based finetuning is more data-efficient --- it outperforms discriminative methods in few-shot scenarios.
\end{abstract}

\section{Introduction}

Inspired by the success of large-scale pre-training model in natural language processing, various vision-language pre-training methods that aim to learn multimodal representations from large-scale image-text pairs have been proposed. Once pretrained, these pre-trained checkpoints can be fine-tuned to perform various downstream tasks. This simple pretrain-and-finetune paradigm has recently shown great potential in many challenging vision-and-language tasks such as visual question answering, image-text retrieval, image captioning and visual entailment\cite{ViLBERT,vlbert,12in1,uniter,Oscar}.

\begin{figure}[t]
\begin{center}
   \includegraphics[scale=0.45]{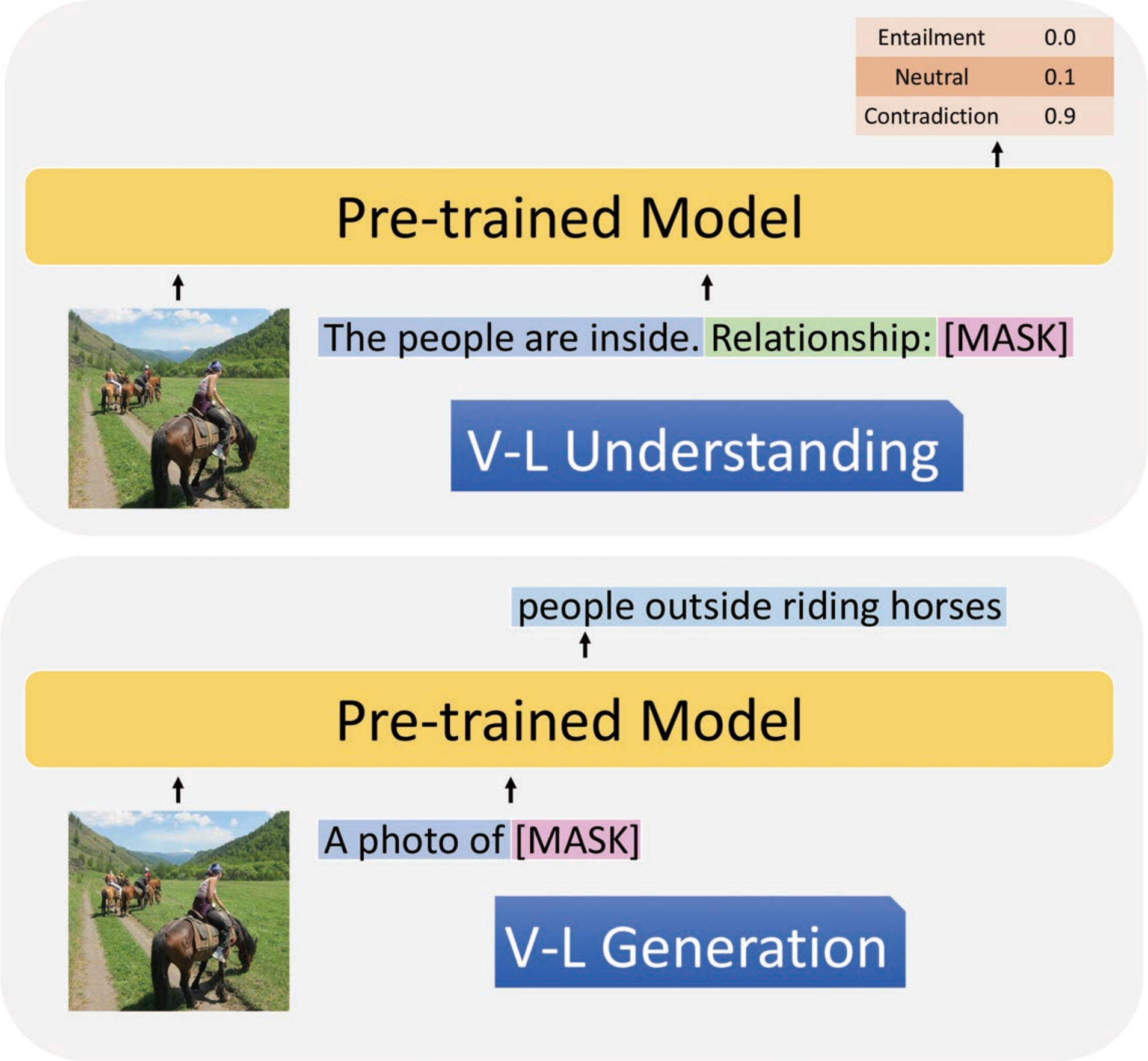}
\end{center}
   \vspace{-0.2in}
   \caption{ We propose a unified pre-trained model that can handle both understanding and generation tasks. }
\label{fig:long}
\label{fig:onecol}
\vspace{-0.2in}
\end{figure}

Vision-language downstream tasks fall into two categories: understanding and generation. The understanding tasks include visual question answering (VQA), visual entailment (VE), image classification and image-text retrieval. Most existing vision-language pre-training methods formulate tasks in this category as discriminative tasks, and they require the model to select an answer from a predefined answer list, e.g. for VQA, existing methods formulate it as a multi-answer classification task and feed the $\rm[CLS]$ output representation to an additional linear classifier\cite{visualbert,12in1,uniter,vilt,villa}. These tasks typically require the model to have a coarse understanding of the semantic information of the image and text, e.g., is the text describing the content of the image? what is the relationship between the image and text? entailment, neutral, or contradictory? In contrast, generation tasks usually require the model to generate a complete sentence conditioned on an image. A typical example is image captioning which requires the model to output a sentence describing the content of the image. 
Existing vision-language methods employ BERT-like objectives, such as masked language modeling and image-text matching\cite{vlbert,vilt,Oscar,uniter} to learn multimodal representations. They perform well on understanding tasks but cannot be directly applied to generative tasks.

We propose Unified Multimodal Pre-training for Vision-Language Understanding and Generation (UniVL) that handles understanding and generative tasks with shared parameters. We first use an image encoder and a text encoder to encode the image and text separately. Then we use a multimodal encoder to fuse the image and text features with cross-attention. Like existing vision-language pretraining methods, two common objectives including masked language modeling and image-text matching are used. However, different from previous methods, we use not only the fully-visible mask but also the causal mask during pre-training. The benefit of the causal mask is to allow the model to do autoregressive decoding, which is important for full sentence generation. Unifying understanding and generation pre-training leads to a unified model with shared parameters alleviating the need to train different models.

Typical vision-language pre-training methods always train multiple task-specific heads for different downstream tasks. This strategy is to adapt the pre-trained model to different downstream tasks and it requires designing task-specific objective functions and separately-optimized architectures\cite{Oscar,vlbert,ViLBERT}. Prompt-based method has recently attracted people's attention and has proven to be simple and effective\cite{gpt3}. Downstream task can be reformulated to the task format that the pre-trained model learned during pre-training. Take topic classification for example, the input is a sentence, ''He is playing basketball.'', the output is a multi-class label, health, politics, sports or others. Instead of adding another linear classifier for fine-tuning the pre-trained model on this task, we can construct a query sentence, ''He is playing basketball. The text is about \_'', and ask the pre-trained model to fill the blank. The filling-blank task is familiar to the pre-trained model, since it is one of the pre-training objectives (masked language modeling). In contrast to the widely used pretrain-and-finetune paradigm that adapts pre-trained model to different downstream tasks, prompt-based methods convert different downstream tasks to the pre-trained task and it can better unleash the potential of the pre-trained model. 
We formulate some previous classification tasks as text generation tasks and use prompt to fine-tune the pre-trained model.

To evaluate the effectiveness of our UniVL, we compare our model with recent vision-language transformers on a diverse set of understanding and generative downstream tasks, including image-text retrieval, visual question answering, visual entailment, image captioning and fine-grained image classification. Our UniVL achieves comparable performance to recent state-of-the-art vision-language pre-training methods on both understanding and generative tasks. We also explore the prompt-based method in our UniVL for different downstream tasks. Our experiments show that the prompt-based method not only reaches comparable performance to previous discriminative methods but also brings other benefits. The prompt-based generative method shows better generalization ability than discriminative method: on VQA, the prompt-based generative method can generate open-ended answers and has a significant improvement on questions with rare answers. The prompt-based method also shows better few-shot learning ability than discriminative methods in different downstream tasks.

The contributions can be summarized as following:
\begin{itemize}
\item We propose Unified multimodal pre-training for Vision-Language understanding and generation (UniVL), which has both understanding and generation ability by mixing causal mask with bidirectional attention mask during pre-training.

\item We propose the prompt-based method to adapt different downstream tasks to the pre-trained model, which outperforms previous discriminative method and shows better generalization ability and few-shot learning ability.

\end{itemize}

\section{Related Work}
\paragraph{Vision-Language Pre-training.} Following the success of large-scale language model pre-training, vision-language pre-training has recently been actively explored and shows superior performance\cite{ViLBERT,vlbert,Oscar,uniter,vilt}. Most of existing methods use the transformer architecture and follow the BERT-like objectives: (i) multimodal masked language modeling\cite{vilt,Oscar,vlbert}: predicting masked words or masked visual features conditioned on the neighboring visual and textual context; (ii) image-text matching\cite{ViLBERT,vilt}: predicting whether the input image is matched with the input text. They use fully-visible mask in the self-attention block, which causes a discrepancy between pretrainining and downstream tasks that require autoregressive generation. Inspired by \cite{unilm,unilmv2}, we mix causal mask with bidirectional attention mask during pre-training such that our model can have both understanding and generation ability by design. \cite{aclunivl} also explores a unified understanding and generation framework for video language modeling with standard fine-tuning approaches. In contrast, we focus on prompt-based fine-tuning strategies for downstream vision-language tasks.

\paragraph{Prompt-based learning in NLP.}
With the increasing number of the parameters of pre-trained models, it is prohibitive to fine-tune all parameters for a downstream task\cite{gpt3,mega}. The idea of prompt methods is to convert the data of downstream tasks into the pattern that is likely already seen during large-scale pretraining, e.g., in NLP, a sentiment classification task can be formulated as a natural language sentence with masked tokens and the models need to fill in the word \textit{positive} or \textit{negative}. Thanks to the similarity of data inputs, the pretrained models can be directly applied to downstream tasks without parameter tuning.
Early prompt-based methods usually adopt hand-crafted templates. For instance \cite{clozeknow} manually designed cloze templates for knowledge probe tasks in the LAMA dataset. \cite{gpt3} designed prefix templates for QA, translation and probe tasks. While these hand-craft prompts are highly interpretable and often effective, they require lots of trials and errors and different heuristics are required for different tasks.
More recently, various methods attempt to automatically learn the optimal prompt. Depending on the search space, these approaches can be divided into discrete and continuous methods. Discrete methods~\cite{pmining,pparaBART,pparaBERTese,pgradientAuto,pgradientUni,pgenPADA,pgen,pscoring} attempt to find a sequence of words as the query prompt that yield the best performance. Instead of explicit searching for the word sequences, continuous methods~\cite{prefix,pinit,pinitWarp,phard,phardPTR} aim to find a sequence of continuous embeddings that can be prepended to the embedded inputs. These hidden prompts do not correspond to any interpretable sentences but can make the search problem much easier.
We experiment with both hand-crafted prompts and continuous prompts.

\section{Approach}

\subsection{Model Architecture}

\paragraph{Visual Encoder.}
We use a ViT~\cite{vit} pre-trained on ImageNet-1k from \cite{deit} as the visual encoder to extract image features. The input image $\textit{I}\in{\mathbb{R}^{C\times{H}\times{W}}}$ is first reshaped into $N=HW/P^2$ flattened 2D patches $v\in{\mathbb{R}^{N\times{(P^2\bigcdot{C})}}}$, where the resolution of the input image is $H\times{W}$, $C$ is the number of channels and the resolution of each patch is $P\times{P}$. Similar to BERT's $\rm [CLS]$ token, ViT prepends a learnable start-of-sequence token to the patch sequence. Using a linear projection $V \in {\mathbb{R}^{(P^{2}{\cdot}C)\times{H}}}$ and the position embedding $V_{pos}\in{\mathbb{R}^{(N+1)\times{H}}}$ layer, $v$ is embedded into $\bar{v}\in{\mathbb{R}^{N\times{H}}}$. The visual encoder consists of alternating layers of multiheaded self-attention (MSA) and MLP blocks, where the MLP contains two layers with a GELU function. Layer Normalization (LN) and residual connection are also used in each layer.
\begin{equation}
    \begin{aligned}
    z_0 &= [v_{CLS};v^1_{p}V;v^2_{p}V;...;v^N_{p}V] + V_{pos} \\ 
    z_{l}' &= \mathbf{MSA}(\mathbf{LN}(z_{l-1})) + z_{l-1}, l=1...L_V \\
    z_l &= \mathbf{MLP}(\mathbf{LN}(z_{l}')) + z_l', l=1...L_V
    \end{aligned}
\end{equation}
where $v^1_{p},...,v^N_{p}$ are flattened 2D patches, $v_{CLS}$ is the learnable embedding, $z_l$ is the hidden states of the $l$ layer.

\paragraph{Text Encoder.}
We use BERT\cite{bert} as the text encoder. The text encoder also consists of several layers of MSA and MLP blocks same as the visual encoder except that LN comes after MSA and MLP. The input text $t\in{\mathbb{R}^{L\times{O}}}$ is embedded to $\Bar{t}\in{\mathbb{R}^{L\times{H}}}$ with a word embedding matrix $T\in{\mathbb{R}^{O\times{H}}}$ and position embedding $T_{pos}\in{\mathbb{R}^{(L+1)\times{H}}}$.

\begin{equation}
    \begin{aligned}
    p_0 &= [t_{CLS};t^1T;t^2T;...;t^NT] + T_{pos} \\ 
    p_{l}' &= \mathbf{LN}(\mathbf{MSA}(p_{l-1})) + p_{l-1}, l=1...L_T \\
    p_l &= \mathbf{LN}(\mathbf{MLP}(p_{l}')) + p_l', l=1...L_T
    \end{aligned}
\end{equation}
where $t^1,...,t^N$ are input words, $p_l$ is the sequence of hidden states of the $l$ layer.

\paragraph{Multimodal Encoder.} The multimodal encoder is similar to the text encoder, except for one more step of cross attention to fuse image features and text features.

\begin{equation}
    \begin{aligned}
    m_0 &= p_{L_T} \\ 
    m_{l}'' &= \mathbf{LN}(\mathbf{MSA}(m_{l-1})) + m_{l-1}, l=1...L_M \\
    m_{l}' &= \mathbf{LN}(\mathbf{MCA}(m_{l}'',z_{L_V})) + m_{l}'', l=1...L_M \\
    m_l &= \mathbf{LN}(\mathbf{MLP}(m_{l}')) + m_{l}', l=1...L_M
    \end{aligned}
\end{equation}
where $p_{L_T}$ is the output of the text encoder, $z_{L_V}$ is the output of the visual encoder, $m_l$ is the sequence of hidden states of the $l$ layer.

\begin{figure}[t]
\begin{center}
   \includegraphics[scale=0.50]{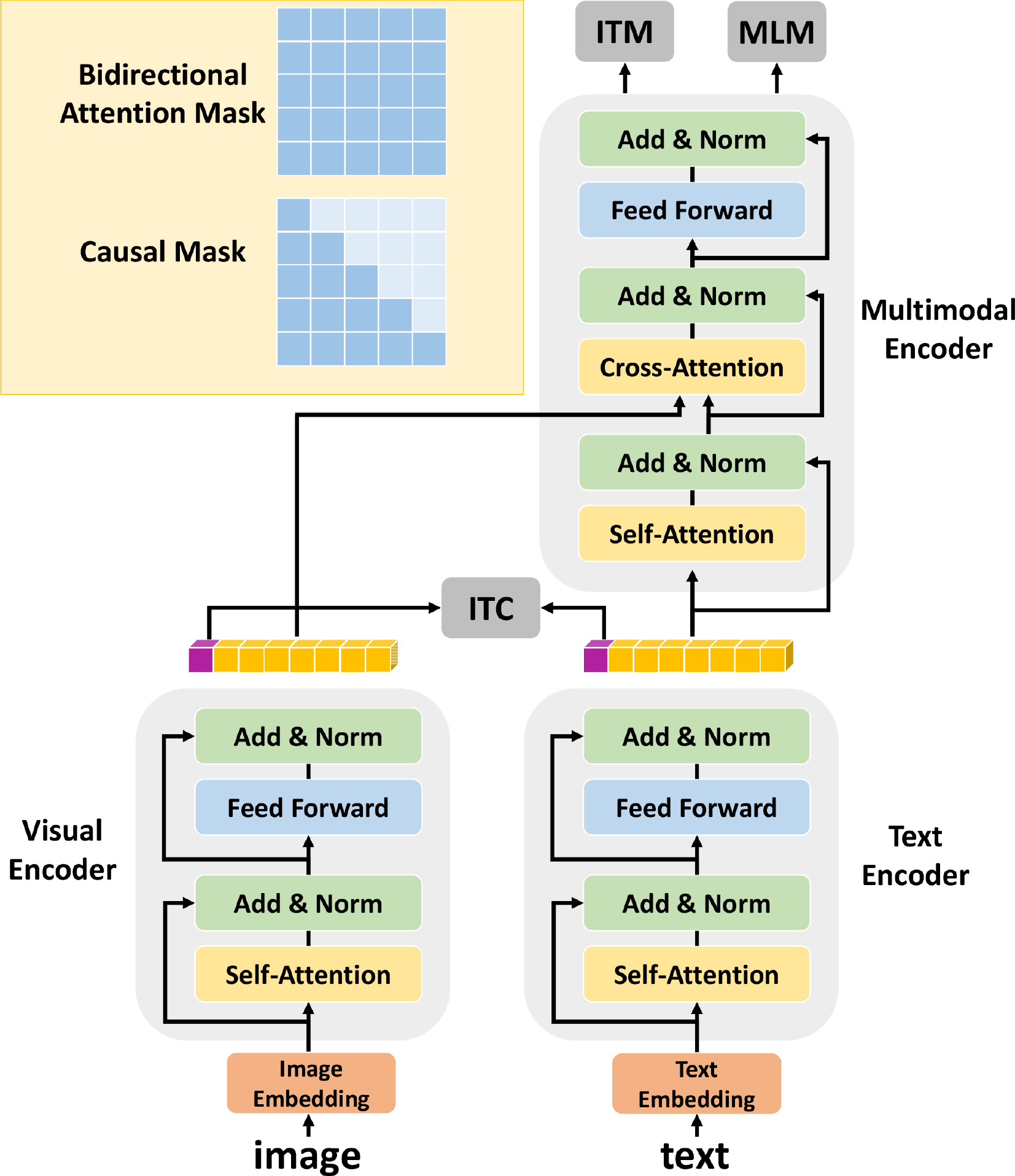}
\end{center}
   \vspace{-0.2in}
   \caption{ Illustration of our UniVL. }
\label{fig:long}
\label{fig:onecol}
\end{figure}

\paragraph{Attention Masks.}
Typical attention mask used in multiheaded self-attention is bi-directional and each token can attend to all tokens. The bi-directional attention mask performs well in discriminative tasks, but it is not suitable for generative tasks. Typically generative tasks require the models to generate tokens in an autoregressive fashion, i.e., from left to right. To accomendate this at pretraining time, we mix the two attention masks in different proportions in the self-attention blocks of the text encoder and the multimodal encoder.

\begin{figure}[t]
   \includegraphics[scale=0.40]{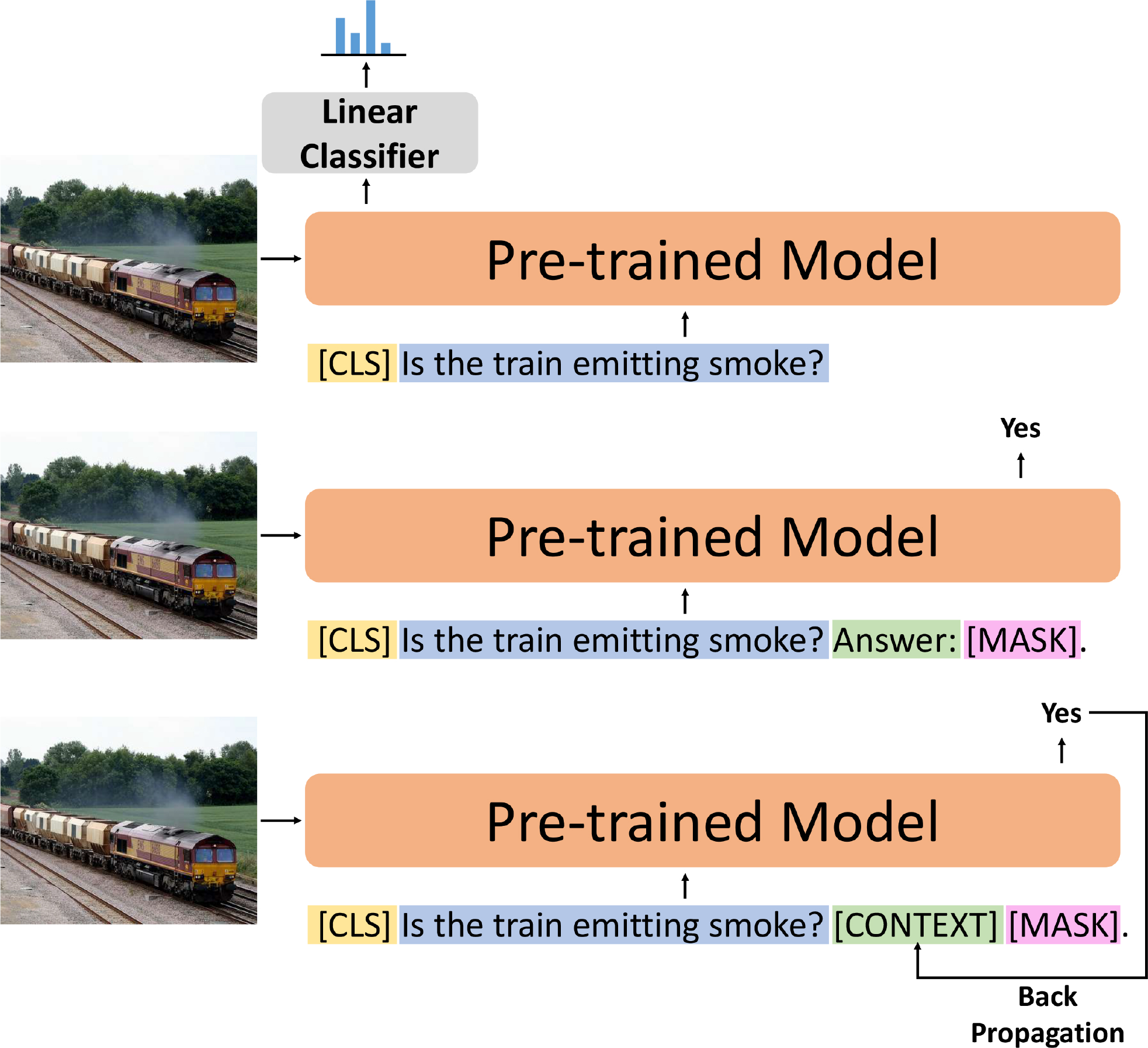}
   \caption{Comparison between previous discriminative methods and prompt-based generative methods.}
\label{fig:long}
\label{fig:onecol}
\label{fig:fig3}
\vspace{-0.2in}
\end{figure}

\subsection{Multimodal Prompting}

Previous vision-language pre-training methods always use the $\rm[CLS]$ token which attends to all image tokens and text tokens as the image-text multimodal representation and add other linear layers to fine-tune for downstream tasks\cite{Oscar,uniter,vilt,vlbert}. For example, in \cite{uniter}, VQA is formulated as a multi-answer classification problem, and it takes the $\rm[CLS]$ token as input to linear layers for fine-tuning. However, we formulate VQA as a text generation problem using prompt. For example, when answering the question ``What is he on top of?'', we can continue with a prompt ``Answer: \_\_\_'', so the whole input of the model is ``What is he on top of? Answer: \_\_\_'' and we ask the pre-trained model to fill the blank. Furthermore, as shown in Figure \ref{fig:fig3}, we replace the language prompt with learnable context. This is because designing a proper natural language prompt is hard, which requires domain expertise and will take an amount of time for tuning since a slight change of each word in the natural language prompt could have a huge impact on downstream task performance. We use the $\rm[unused]$ tokens of the tokenizer as the learnable contexts since they are parameterized and can be updated via back propagation.

\subsection{Pre-training Objectives}
\paragraph{Image-Text Contrastive Loss.}
The image-text contrastive loss has proven to be effective for vision-language pre-training\cite{clip}. We apply image-text contrastive loss to learn a common low-dimension space to embed images and texts. We treat matched image-text pairs as positive and all other random image-text pairs in the training batch as negative. We minimize the sum of two losses: one for image-to-text and one for text-to-image.

\begin{equation}
    \begin{aligned}
    \mathcal{L}_{i2t} &= -\frac{1}{B}\sum^B_ilog\frac{\exp(x_i^Ty_i/\sigma)}{\sum^B_{j=1}\exp(x_i^Ty_i/\sigma)}
    \\ 
    \mathcal{L}_{t2i} &= -\frac{1}{B}\sum^B_ilog\frac{\exp(y_i^Tx_i/\sigma)}{\sum^B_{j=1}\exp(y_i^Tx_i/\sigma)}
    \\
    \mathcal{L}_{itc} &= \mathcal{L}_{i2t} + \mathcal{L}_{t2i}
    \end{aligned}
\end{equation}
where $x_i$ is the first hidden state of the visual encoder output and $y_i$ is the first hidden state of the text encoder. $B$ is the batch size, and $\sigma$ is the temperature to scale the logits. 

\paragraph{Masked Language Modeling.} We randomly mask out the text tokens with a probability of 15\% and replace them with a special $\rm [MASK]$ token and the model needs to predict the masked words using the image and contextual text.

\begin{equation}
    \begin{aligned}
    \mathcal{L}_{mlm} = \sum {\rm H}(p_{mask},y_{mask})
    \end{aligned}
\end{equation}
where $\rm{H}$ is cross-entropy, $p_{mask}$ denotes the model's predicted probability for a masked token and $y_{mask}$ is a one-hot vocabulary distribution where the ground-truth token has a probability of 1, the $\mathcal{L}_{mlm}$ is the sum of each masked word's cross-entropy.

\paragraph{Image-Text Matching.} We use the first token of the multimodal encoder's output as the fused representation of both modalities, and append a fully-connected layer followed by softmax to predict a two-class probability $p_{itm}$. The image-text matching loss predicts whether a pair of image and text is matched (positive) or not matched (negative). The negative pair is created by replacing the image or text in a matched sample with a randomly-selected one from other samples.

\begin{equation}
    \begin{aligned}
    \mathcal{L}_{itm} = \sum {\rm H}(p_{itm},y_{itm}) 
    \end{aligned}
\end{equation}
where $y_{itm}$ is a two-dimensional one-hot vector representing the ground-truth label (1 for matched image-text pairs and 0 for not matched pairs), $\mathcal{L}_{itm}$ is the sum of cross-entropy of all positive and negative samples.
\begin{table*}[t!]
\centering
\begin{tabular}{lccllllll}
\toprule
\multicolumn{1}{c}{\multirow{3}{*}{Method}} & \multirow{3}{*}{\begin{tabular}[c]{@{}c@{}}Number of\\ Pre-train\\ Images\end{tabular}} && \multicolumn{6}{c}{\begin{tabular}[c]{@{}c@{}}Multimodal Retrieval Results on Flickr30K\\ (Zero-shot/Fine-tuned)\end{tabular}}                                                                \\ \cmidrule{4-9} 
\multicolumn{1}{c}{}                        &                                                                                         & \multicolumn{3}{c}{\qquad\qquad image-to-text}                                                               & \multicolumn{3}{c}{\qquad\qquad text-to-image}                                                          \\ \cmidrule{4-9} 
\multicolumn{1}{c}{}                        &                                                                                         && \multicolumn{1}{c}{R@1}       & \multicolumn{1}{c}{R@5}       & \multicolumn{1}{c}{R@10}      & \multicolumn{1}{c}{R@1}       & \multicolumn{1}{c}{R@5}       & \multicolumn{1}{c}{R@10} \\ \cmidrule{1-2}\cmidrule{4-9}
UNITER\cite{uniter}                                        & 4M                                                                                      && \multicolumn{1}{l}{83.6/87.3} & \multicolumn{1}{l}{95.7/98.0} & \multicolumn{1}{l}{97.7/99.2} & \multicolumn{1}{l}{68.7/75.6} & \multicolumn{1}{l}{89.2/94.1} & 93.9/96.8                 \\ 
CLIP\cite{clip}                                          & 400M                                                                                    && \multicolumn{1}{l}{88/-}      & \multicolumn{1}{l}{98.7/-}    & \multicolumn{1}{l}{99.4/-}    & \multicolumn{1}{l}{68.7/-}    & \multicolumn{1}{l}{90.6/-}    & 95.2/-                    \\ 
ALIGN\cite{align}                                         & 1.2B                                                                                    && \multicolumn{1}{l}{\textbf{88.6}/\textbf{95.3}} & \multicolumn{1}{l}{98.7/\textbf{99.8}} & \multicolumn{1}{l}{99.7/\textbf{100}}  & \multicolumn{1}{l}{\textbf{75.7}/\textbf{84.9}} & \multicolumn{1}{l}{\textbf{93.8}/\textbf{97.4}} & \textbf{96.8}/\textbf{98.6}                 \\ \cmidrule{1-2}\cmidrule{4-9}
UniVL                                    & 3M                                                                                        && \multicolumn{1}{l}{86.8/94.3} & \multicolumn{1}{l}{\textbf{98.7}/99.4} & \multicolumn{1}{l}{\textbf{99.7}/99.8} & \multicolumn{1}{l}{73.4/82.8} & \multicolumn{1}{l}{92.1/96.7} & 96.0/98.4                 \\ \bottomrule
\end{tabular}
\vspace{-0.1in}
\caption{Zero-shot and fine-tuned image-to-text retrieval results on Flickr30K}
\label{tab:retrieval}
\vspace{-0.2in}
\end{table*}

\section{Experiments}

\subsection{Pre-training Setup}
\paragraph{Datasets.}
We construct our pre-training data using Conceptual Captions\cite{gcc} and COCO\cite{coco}. We follow the Karpathy's split of COCO\cite{kar} and the total number of unique images is 113K. After filtering out the unavailable urls, we collect 2.84M unique images in total.

\vspace{-0.05in}
\paragraph{Implementation Details.}
We use a 12-layer vision transformer ViT-B/16 as the image encoder, which is initialized with weights pre-trained on ImageNet-1k. The text encoder is initialized using the first six layers of the BERT model and the multimodal encoder is initialized using the last six layers of the BERT model. We pre-train our model for 30 epochs using a batch size of 2048 on 32 NVIDIA Tesla V100 GPUs. We use AdamW optimizer with a learning rate of $1e^{-4}$ and weight decay of 0.02.

\subsection{Downstream Tasks}
\textbf{Image-Text Retrieval} is the task of identifying an image from candidates given a caption describing its content, or vice versa. So it contains two subtasks: image-to-text retrieval and text-to-image retrieval. We use the Image-Text Contrastive Loss and Image-Text Matching Loss to evaluate the similarities between an image and a text. During inference, we first compute feature similarities using the visual encoder and text encoder for all image-text pairs. Then we select top-k pairs and use the multimodal encoder to calculate the image-text matching scores for ranking. We evaluate our model on Flickr30K\cite{flickr} and COCO.

\textbf{Image Captioning} aims to generate a sentence describing an image's content. Since we pre-train our model using causal mask, our model can directly generate a sentence for an image. During sentence generation, we first encode the image and the input of the text encoder with $\rm{[CLS] [MASK]}$. The special token $\rm{[CLS]}$ is the beginning of a sentence and our model predicts the word in the $\rm{[MASK]}$ position. Then we append another $\rm{[MASK]}$ token to the generated token sequence and predict another word. The generation process terminates when the model outputs $\rm{[SEP]}$. We use beam search (beam size = 5) in our experiments and report the results on COCO image captioning dataset.

\textbf{Visual Question Answering} requires the model to answer a question given an image. Existing methods often formulate VQA as a multi-answer classification problem. We consider VQA as a text generation task. We design two prompts for VQA: natural language prompt and learnable context prompt. We evaluate our methods on the VQAv2.

\textbf{Fine-grained Image Classification} focuses on differentiating between hard-to-distinguish object classes, such as species of flowers, or animals. We use the fine-grained image classification task to evaluate the multimodal understanding ability of our model. We also formulate the image classification task as text generation and design natural language prompt and learnable context prompt. Compared with discriminative methods, prompt-based method has better few-shot learning ability. We evaluate our method on Food101\cite{food101}, Flowers102\cite{flowers102}, DTD\cite{dtd}.

\textbf{Visual Entailment}\cite{ve} is a fine-grained visual reasoning task to predict whether the relationship between an image and a text is entailment, neutral, or contradictory. We design natural language prompt and learnable context prompt for VE, and compare them with discriminative methods.

\section{Main Results}

\subsection{Image-Text Retrieval}
We use the image-text retrieval task to evaluate the vision-language understanding ability of the pre-trained model. Table \ref{tab:retrieval} reports the results of zero-shot and fine-tuned image-text retrieval on Flickr30K. For zero-shot retrieval, UniVL reaches comparable results to CLIP and ALIGN with much fewer data. For fine-tuned retrieval, recalls for UniVL are higher by a large margin than UNITER and are similar to ALIGN despite its pre-training on a larger dataset (1.2B).

\subsection{Image Captioning}
We use the image captioning task to evaluate the generation ability of our pre-trained model. We evaluate automatic caption generation performance on COCO dataset, following Karpathy's split\cite{kar}, which re-splits training images and validation images into 113287, 5000, 5000 for training, validation, testing respectively. As shown in Table \ref{tab:caption}, we report results on four popular metrics: BLEU@4\cite{b4}, CIDEr\cite{cider}, METEOR\cite{meteor}, SPICE\cite{spice}. We compare our pre-trained model with other generative vision-language pre-training methods. Our unified model achives comparable performance to recent generative pre-training methods.

\begin{table}[h!]
\centering
\begin{tabular}{lrrrrr}
\toprule
\multicolumn{1}{c}{Method} && \multicolumn{1}{c}{B} & \multicolumn{1}{c}{C} & \multicolumn{1}{c}{M} & \multicolumn{1}{c}{S} \\ \cmidrule{1-1} \cmidrule{3-6} 
Unified VLP\cite{unifiedvlp} &                 & 36.5                   & 117.7                  & 28.4                   & 21.3                   \\ 
XGPT\cite{xgpt}    &                     & \textbf{37.2}                   & \textbf{120.1}                  & 28.6                   & 21.8                   \\ 
VL-T5\cite{vlbart}    &                    & 34.6                   & 116.1                  & \textbf{28.8}                   & \textbf{21.9}                   \\ 
VL-BART\cite{vlbart}   &                   & 34.2                   & 114.1                  & 28.4                   & 21.3                   \\ \cmidrule{1-1} \cmidrule{3-6}
UniVL     &              & 35.6                   & 116.8                  & 28.6                   & 21.4                   \\ \bottomrule
\end{tabular}
\vspace{-0.1in}
\caption{Image captioning scores on Karpathy's test split\cite{kar}.}
\label{tab:caption}
\vspace{-0.2in}
\end{table}

\subsection{Visual Question Answering}
We formulate VQA as a text generation task instead of a multi-answer classification task. We design natural language prompt and learnable context prompt for VQA. The prompt is ``$[\rm QUESTION]$ \xspace Answer: \xspace $\rm[ANSWER]$'', where $\rm[QUESTION]$ indicates the question text and $\rm[ANSWER]$ indicates the answer text. We mask out the tokens of $\rm[ANSWER]$ and optimize the masked language modeling loss during fine-tuning. During inference, the input text is ``$[\rm QUESTION]$  \xspace Answer: \xspace $\rm[MASK]$''. The model predicts the word and appends $\rm[MASK]$ to the generated sequence iteratively until generating $\rm[SEP]$ token. The learnable context prompt replaces the natural language prompt with learnable tokens, and for VQA it is \textit{``$\rm [QUESTION]$ $\rm[CTX]$ $\rm[ANSWER]$''}. $\rm [CTX]$ is the sequence of learnable tokens and we use the $\rm[unused]$ tokens of BERT tokenizer as $\rm[CTX]$. The length of $\rm[CTX]$ is 16. Compared with natural language prompt, $\rm[CTX]$ is a parameterized prompt and can be updated along with other parameters.

Previous discriminative methods formulate VQA as a multi-answer classification problem and the model is asked to select an answer from a predefined answer list. To get a fine-grained comparison of discriminative methods and generative methods, we divide the Karpathy's test split\cite{kar} into two categories: questions whose answer in the predefined answer list (in-domain samples) and questions whose answer not in the list (out-domain samples). The size of the predefined answer list is 3129, and the number of in-domain samples and out-domain samples is 25750 and 530. Typical discriminative methods cannot answer the out-domain questions, since their answers are rare and not in the predefined answer list. To compare the generalization ability of discriminative and generative methods, we append answers of out-domain samples to the predefined answer list and use the expanded answer list to fine-tune discriminative method. For discriminative method, we feed the first hidden state of the multimodal encoder's output into an additional linear classifier for answer prediction. As shown in Table \ref{tab:vqa}, compared with discriminative method, prompt-based generative methods perform better in both categories and the improvement is more significant when comparing them in the out-domain samples, showing better generalization ability with prompting methods. To evaluate the few-shot learning ability of prompt-based methods, we use different amounts of the Karpathy's training data\cite{kar}. As shown in Table \ref{tab:vqafew}, both natural language prompt and learnable context prompt outperform the discriminative methods.

\begin{table}[t]
\centering
\begin{tabular}{lclll}
\toprule
Method && in-domain & out-domain & overall \\ \cmidrule{1-1} \cmidrule{3-5}
LC     && 70.8      & 3.7        & 69.4    \\ 
NLP    && 68.4      & 13.9       & 67.3    \\ 
LCP    && \textbf{72.1}      & \textbf{15.1}       & \textbf{71.0}    \\ \bottomrule
\end{tabular}
\vspace{-0.1in}
\caption{VQA results of discriminative method and generative methods on Karpathy's test split\cite{kar}. LC, NLP and LCP means linear classifier, natural language prompt and learnable context prompt.}
\label{tab:vqa}
\vspace{-0.1in}
\end{table}

\begin{table}[t!]
\centering
\scalebox{0.9}{
\begin{tabular}{cccccc}
\toprule
\multirow{2}{*}{Method} && \multicolumn{4}{c}{\begin{tabular}[c]{@{}c@{}}Number of training samples\\ (in-domain/out-domain)\end{tabular}} \\ \cmidrule{3-6} 
                        && \multicolumn{1}{c}{4K}         & \multicolumn{1}{c}{22K}        & \multicolumn{1}{c}{44K}        & 88K        \\ \cmidrule{1-1}\cmidrule{3-6}
LC                      && \multicolumn{1}{c}{0.5/0}      & \multicolumn{1}{c}{5.6/0}      & \multicolumn{1}{c}{10.9/0.5}   & 15.4/0.9   \\ 
NLP                     && \multicolumn{1}{c}{0.9/\textbf{0.1}}    & \multicolumn{1}{c}{11.9/\textbf{0.9}}   & \multicolumn{1}{c}{14.8/1.1}   & 18.3/1.6   \\ 
LCP                     && \multicolumn{1}{c}{\textbf{0.9}/0}      & \multicolumn{1}{c}{\textbf{12.4}/0.7}   & \multicolumn{1}{c}{\textbf{16.7}/\textbf{1.5}}   & \textbf{20.1}/\textbf{2.4}   \\ \bottomrule
\end{tabular}}
\vspace{-0.1in}
\caption{VQA results of different numbers of training samples on different methods.}
\label{tab:vqafew}
\vspace{-0.2in}
\end{table}

To get a fair comparison with recent state-of-the-art vision-language pre-training methods, following previous methods\cite{uniter,LXMERT}, we use the VQAv2 training and validation set and additional training samples from Visual Genome to fine-tune the pre-trained model. As shown in Table \ref{tab:vqavesota}, our UniVL reaches comparable performance to state-of-the-art methods.

\begin{table}[b]
\centering
\scalebox{0.9}{
\begin{tabular}{cccccc}
\toprule
\multirow{2}{*}{Method} && \multicolumn{2}{c}{VQA}                 & \multicolumn{2}{c}{VE}            \\ \cmidrule{3-6} 
                        && \multicolumn{1}{c}{test-dev} & test-std & \multicolumn{1}{c}{val}   & test  \\ \cmidrule{1-1}\cmidrule{3-6} 
VisualBERT\cite{visualbert}              && \multicolumn{1}{c}{70.8}     & 71       & \multicolumn{1}{c}{-}     & -     \\ 
12-in-1\cite{12in1}                 && \multicolumn{1}{c}{73.15}    & -        & \multicolumn{1}{c}{-}     & 76.95 \\ 
UNITER\cite{uniter}                  && \multicolumn{1}{c}{72.7}     & 72.91    & \multicolumn{1}{c}{78.59} & 78.28 \\ 
VilT\cite{vilt}                    && \multicolumn{1}{c}{70.94}    & -        & \multicolumn{1}{c}{-}     & -     \\ 
VILLA\cite{villa}                   && \multicolumn{1}{c}{\textbf{73.59}}    & \textbf{73.67}    & \multicolumn{1}{c}{79.47} & 79.03 \\ \cmidrule{1-1}\cmidrule{3-6} 
UniVL                   && \multicolumn{1}{c}{72.31}    & 72.53    & \multicolumn{1}{c}{\textbf{79.70}} & \textbf{80.00} \\ \bottomrule
\end{tabular}}
\vspace{-0.1in}
\caption{Comparison with state-of-the-art vision-language pre-training methods on VQA and VE.}
\label{tab:vqavesota}
\vspace{-0.2in}
\end{table}

\subsubsection{Fine-grained Image Classification}
Similar to VQA, we formulate image classification as a text generation task and design natural language prompt and learnable context prompt for image classification. The natural language prompt is $\rm a\ photo\ of\ +\ [CATEGORY]$ and the corresponding learnable context prompt is $\rm\ [CTX]\ +\ [CATEGORY]$. We mask out $\rm[CATEGORY]$ during fine-tuning. The $\rm[CATEGORY]$ is the class name of the image (\eg, cannoli for food classification, and gazania for flower classification). As shown in Table \ref{tab:cls}, learnable context prompt is an effective method for this downstream task, since prompt is a form closer to the pre-training task than the discriminative method.

\begin{table}[h]
\centering
\scalebox{0.9}{
\begin{tabular}{cccccc}
\toprule
Method                     & \begin{tabular}[c]{@{}c@{}}Fine-tuned\\ Component\end{tabular} && Food101 & Flowers102 & DTD \\ \cmidrule{1-1}\cmidrule{2-2}\cmidrule{4-6}
LC                   & VE                                                             && 92.8    & \textbf{93.8}          & \textbf{65.4}   \\ \cmidrule{1-1}\cmidrule{2-2}\cmidrule{4-6}
NLP                  & VE                                                             && 88.4    & 90.1          & 53.3   \\ \cmidrule{1-1}\cmidrule{2-2}\cmidrule{4-6}
\multirow{5}{*}{LCP} & VE                                                             && 92.8    & 93.4          & 62.1   \\ 
                     & TE                                                             && 78.6    & 74.6          & 19.2   \\ 
                     & ME                                                             && 79.4    & 76.2          & 20.6   \\ 
                     & VETE                                                           && 92.8    & 93.5          & 63.3   \\ 
                     & VEME                                                           && \textbf{93.3}    & 93.7          & 63.5   \\ \bottomrule
\end{tabular}}
\vspace{-0.1in}
\caption{Image classification results of discriminative method and generative methods. LC, NLP and LCP means linear classifier, natural language prompt and learnable context prompt. VE, TE, ME means visual encoder, text encoder and multimodal encoder.}
\label{tab:cls}
\vspace{-0.1in}
\end{table}

\begin{figure}[h]
   \includegraphics[scale=0.37]{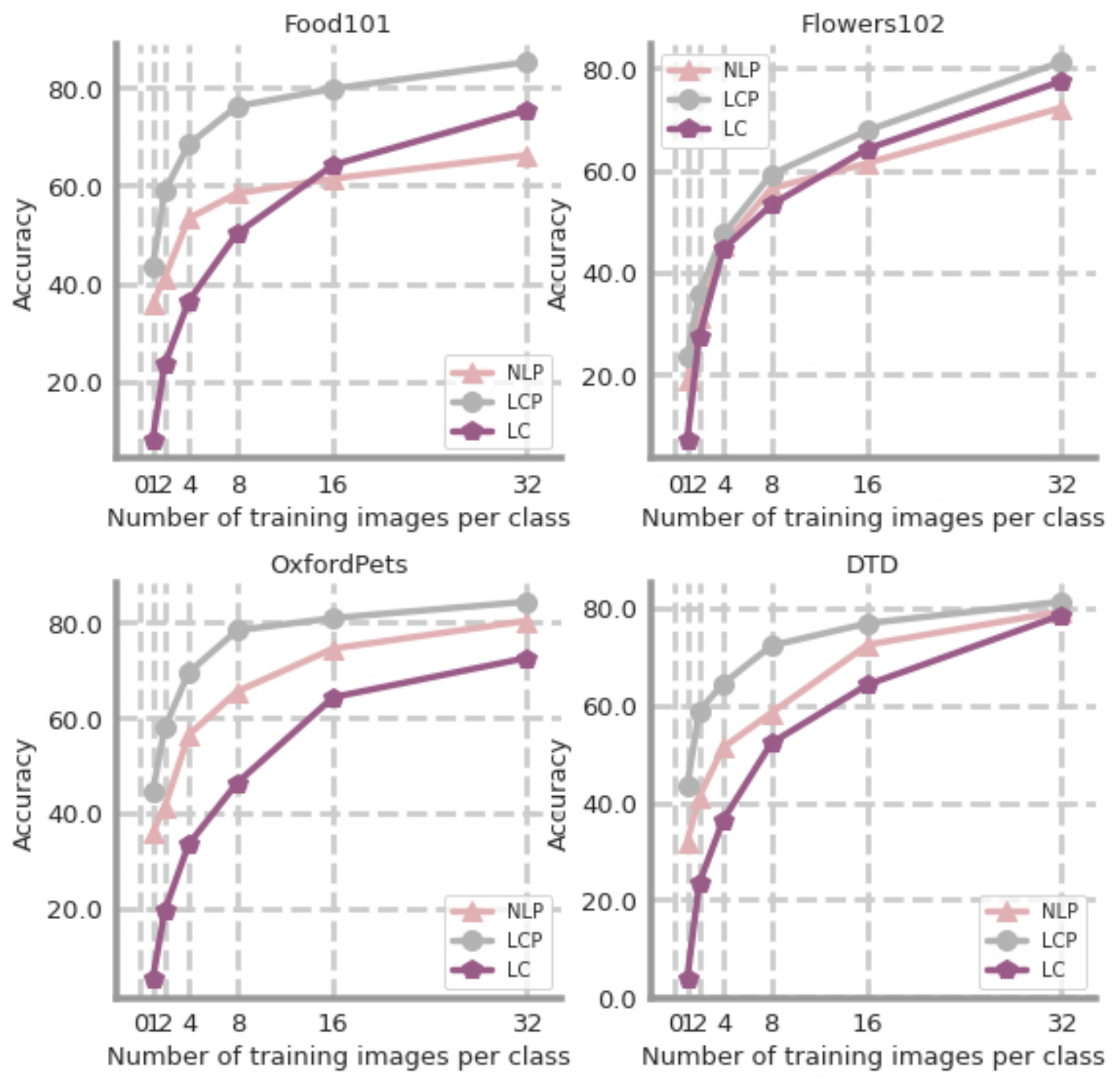}
\vspace{-0.1in}
   \caption{Few-shot learning ability of different methods. }
\label{fig:clsfew}
\end{figure}

Figure \ref{fig:clsfew} shows that the learnable context prompt has better few-shot learning ability. Since prompt is a more familiar task form to the pre-trained model, the learnable context prompt can make better use of the knowledge learned from the pre-trained model. Different from VQA, the image classification task has simper text input and the number of training samples is quite lower, so updating all parameters of the pre-trained model is unwise. As shown in Table \ref{tab:cls}, the visual encoder is the key for the fine-grained image classification task, since the text is simple and contains few semantic information except the class name.

\begin{table}[h]
\centering
\begin{tabular}{cccc}
\toprule
Method    && val  & test \\ \cmidrule{1-1}\cmidrule{3-4}
LC        && 78.4 & 78.1 \\ 
NLP       && 65.4 & 65.7 \\ 
NLP(1in3) && 75.3 & 75.9 \\ 
LCP       && 77.6 & 78.0 \\ 
LCP(1in3) && \textbf{79.7} & \textbf{80.0} \\ \bottomrule
\end{tabular}
\vspace{-0.1in}
\caption{Visual entailment results of different methods.}
\label{tab:ve}
\vspace{-0.2in}
\end{table}

\begin{table}[]
\centering
\scalebox{0.90}{
\begin{tabular}{cccrrrrr}
\toprule
\multirow{2}{*}{\begin{tabular}[c]{@{}c@{}}Downstream\\ Task\end{tabular}} & \multirow{2}{*}{\begin{tabular}[c]{@{}c@{}}Prompt\\ Position\end{tabular}} && \multicolumn{5}{c}{Prompt Length}                                                                                                      \\ \cmidrule{4-8} 
                                                                           &                                                                            && \multicolumn{1}{c}{1}    & \multicolumn{1}{c}{4}    & \multicolumn{1}{c}{8}    & \multicolumn{1}{c}{16}   & \multicolumn{1}{c}{32} \\ \cmidrule{1-2}\cmidrule{4-8}
VQA                                                                        & begin                                                                      && \multicolumn{1}{r}{66.4} & \multicolumn{1}{r}{69.2} & \multicolumn{1}{r}{70.4} & \multicolumn{1}{r}{70.8} & 71.0                    \\ 
VQA                                                                        & mid                                                                        && \multicolumn{1}{r}{67.9} & \multicolumn{1}{r}{69.5} & \multicolumn{1}{r}{70.4} & \multicolumn{1}{r}{71.0}   & 71.1                    \\ 
VE                                                                         & begin                                                                      && \multicolumn{1}{r}{77.3} & \multicolumn{1}{r}{78.2} & \multicolumn{1}{r}{78.8} & \multicolumn{1}{r}{79.4} & 79.9                    \\ 
VE                                                                         & mid                                                                        && \multicolumn{1}{r}{77.5} & \multicolumn{1}{r}{78.5} & \multicolumn{1}{r}{78.6} & \multicolumn{1}{r}{80.1} & 80.1                    \\ 
\begin{tabular}[c]{@{}c@{}}IC(Fo101)\end{tabular}                     & begin                                                                      && \multicolumn{1}{r}{90.6} & \multicolumn{1}{r}{91.4} & \multicolumn{1}{r}{91.9} & \multicolumn{1}{r}{92.1} & 92.5                    \\ 
\begin{tabular}[c]{@{}c@{}}IC(Fl102)\end{tabular}                  & begin                                                                      && \multicolumn{1}{r}{93.0}   & \multicolumn{1}{r}{93.6} & \multicolumn{1}{r}{94.2} & \multicolumn{1}{r}{94.4} & 94.0                      \\ \bottomrule
\end{tabular}}
\vspace{-0.1in}
\caption{Ablation study of different prompt length and prompt position. Fo101 and Fl102 mean Food101 dataset and Flowers102 dataset.}
\label{tab:lenpos}
\vspace{-0.3in}
\end{table}

\input{large_tbl.tex}

\subsubsection{Visual Entailment}
Not all generative methods perform better than discriminative methods. The discriminative methods are more appropriate for some downstream tasks, and visual entailment is one of them. We also design natural language prompt and learnable context prompt by formulating VE as a text generation task. The natural language prompt is $\rm [SENTENCE]\ +\ 'Relationship:'\ +\ [LABEL]$ and the learnable context prompt is $\rm\ [SENTENCE]\ +\ [CTX]\ +\ [LABEL]$. The $\rm [LABEL]$ is the relationship between the image and the $\rm[SENTENCE]$ and it is one of entailment, neutral and contradiction. We mask out $\rm[LABEL]$ during fine-tuning. Different from predicting a word from the vocabulary according to each word's score, we also rank the score of each possible answer and return the top during inference. It is a discriminative method, and compared with using an additional linear classifier, it is more familiar to the pre-trained model, since the input is an image with a text and the objective is masked language modeling. As shown in Table \ref{tab:ve}, 1in3 means that we predict the $\rm[MASK]$ in a set of three words: entailment, neutral and contradiction. It should be noted that we can do this in visual entailment because each answer of visual entailment is a single word thus there is no common prefix, for VQA or image classification, there exist some common prefixes between answers (\eg, ``bus'' and ``bus driver'' for VQA, ``chicken curry'' and ``chicken wings'' for Food101). Compared with generative methods, the discriminative methods are more suitable for visual entailment, since the answer set is too small.

\subsubsection{Ablation Study}
\textbf{Effectiveness of causal mask during pre-training.} We first evaluate the effectiveness of the causal mask during pre-training. Table \ref{tab:p} shows the multimodal generation ability and understanding ability of the pre-trained model with different mixing ratios of bi-directional attention mask and causal mask and different numbers of the image-text pairs during pre-training. We use the image captioning task to evaluate the generation ability of the model, and image classification, visual entailment and image-text retrieval to evaluate the understanding ability. For image captioning, we append the special token $\rm[MASK]$ to the sequence and predict the word iteratively until the model outputs the special token $\rm[SEP]$. For understanding tasks, we feed the first hidden state of the multimodal encoder's output to an additional linear classifier for predicting the answer. The understanding task requires the model to judge the relationship between an image and a sentence, which is a closed task and the model needs to select an answer from a predefined set. The generation task is more difficult as the model needs to generate open-ended answers.

As shown in Table \ref{tab:p}, with the increase of causal mask during pre-training, the model performs better in generative tasks. However, as causal mask increases, bi-mask decreases and the model performs worse in understanding tasks. We found that generally causal mask is beneficial for generative tasks and bi-mask is beneficial for understanding tasks, while the increase of training data is of greater benefit to both understanding and generation tasks. Hence it seems to be a tradeoff between understanding tasks and generation tasks since it hard to design a model that achieves the optimal performances for both tasks at the same time.

\textbf{Prompt position and length.}
We use the $\rm[unused]$ tokens of the $\rm bert-based-uncased$ tokenizer as the learnable context. For VQA and VE, the input text contains a sentence, a prompt and a $\rm[MASK]$ token, and the prompt can be at the beginning of the input text (e.g. $\rm\ [CTX]\ +\ [QUESTION]\ +\ [MASK]$), or in the middle of the input text (e.g. $\rm [QUESTION]\ +\ [CTX]\ +\ [MASK]$). It should be noted that the prompt should not be at the end of the input, since we use causal mask and the $\rm[MASK]$ token can not attend to tokens on the right. For image classification, the input text contains only $\rm[CTX]$ and $\rm[MASK]$, and the prompt should be on the left of $\rm[MASK]$. As shown in Table \ref{tab:lenpos}, we found that only a single one $\rm[CTX]$ is unable to effectively prompt the model, and with the increase of the prompt length, the accuracy of different downstream tasks can be benefited, while too long prompt is inefficient and compared with the length of 16, the length of 32 is doubled, but the accuracy of downstream tasks remains almost unchanged. For VQA and VE, the prompt in the middle outperforms the prompt at the beginning, since the prompt in the middle is closer to the $\rm[MASK]$ token and is a more effective signal for the following generation.

\section{Conclusion}
In this paper we introduced a unified model named UniVL that can handle vision-language understanding and generation tasks and reaches comparable performance to SOTA on understanding and generation tasks. We also proposed prompt-based method, which is a simple and effective method for fine-tuning on different downstream tasks.

\section*{Acknowledgment}
This work is partially supported by Tianshu AI Platform, Zhejiang Lab.

{
\small

\input{PaperForReview.bbl}
\bibliographystyle{ieee_fullname}
}

\section{Appendix}
\begin{table}[]
\scalebox{0.85}{
\begin{tabular}{cccccc}
\toprule
Task                 && Dataset           && \#Image & \#Text \\ \cmidrule{1-1}\cmidrule{3-3}\cmidrule{5-6}
Pre-training         && GCC\cite{gcc}               && 2.84M   & 2.84M  \\ 
                     && COCO\cite{coco}(k-train*)    && 113K    & 567K   \\ \cmidrule{1-1}\cmidrule{3-3}\cmidrule{5-6}
ITR && Flickr30K\cite{flickr}(train)  && 30K     & 150K   \\ 
                     && Flickr30K\cite{flickr}(test)   && 1K      & 5K     \\ \cmidrule{1-1}\cmidrule{3-3}\cmidrule{5-6}
IC     && COCO\cite{coco}(k-test*)     && 5K      & 25K    \\ \cmidrule{1-1}\cmidrule{3-3}\cmidrule{5-6}
VQA                  && VG\cite{vg}                && 99K     & 1.44M  \\ 
                     && VQAv2\cite{vqav2}(train)      && 123K    & 658K   \\ 
                     && VQAv2\cite{vqav2}(test)       && 81K     & 448K   \\ \cmidrule{1-1}\cmidrule{3-3}\cmidrule{5-6}
FIC && Food101\cite{food101}(train)    && 75K     & 75K    \\ 
                     && Food101\cite{food101}(test)     && 25K     & 25K    \\ \cmidrule{3-3}\cmidrule{5-6}
                     && Flowers102\cite{flowers102}(train) && 2K      & 2K     \\ 
                     && Flowers102\cite{flowers102}(test)  && 6K      & 6K     \\ \cmidrule{3-3}\cmidrule{5-6}
                     && DTD\cite{dtd}(train)        && 3K      & 3K     \\ 
                     && DTD\cite{dtd}(test)         && 1K      & 1K     \\ \cmidrule{1-1}\cmidrule{3-3}\cmidrule{5-6}
VE    && SNLI-VE\cite{ve}(train)    && 30K     & 547K   \\ 
                     && SNLI-VE\cite{ve}(test)     && 1K      & 18K    \\ \bottomrule
\end{tabular}
}
\caption{Statistics of datasets. '*' denotes Karpathy's split\cite{kar}. ITR, IC, VQA, FIC, VE mean Image-Text Retrieval, Image Captioning, Visual Question Answering, Fine-grained Image Classification, Visual Entailment.}
\label{tab:dataset}
\end{table}

\section{Dataset Statistics}
We summarize the train/test image and text numbers of our pre-training and downstream datasets in Table \ref{tab:dataset}.

\section{Downstream Tasks}
\begin{figure*}[t]
\begin{center}
   \includegraphics[scale=0.65]{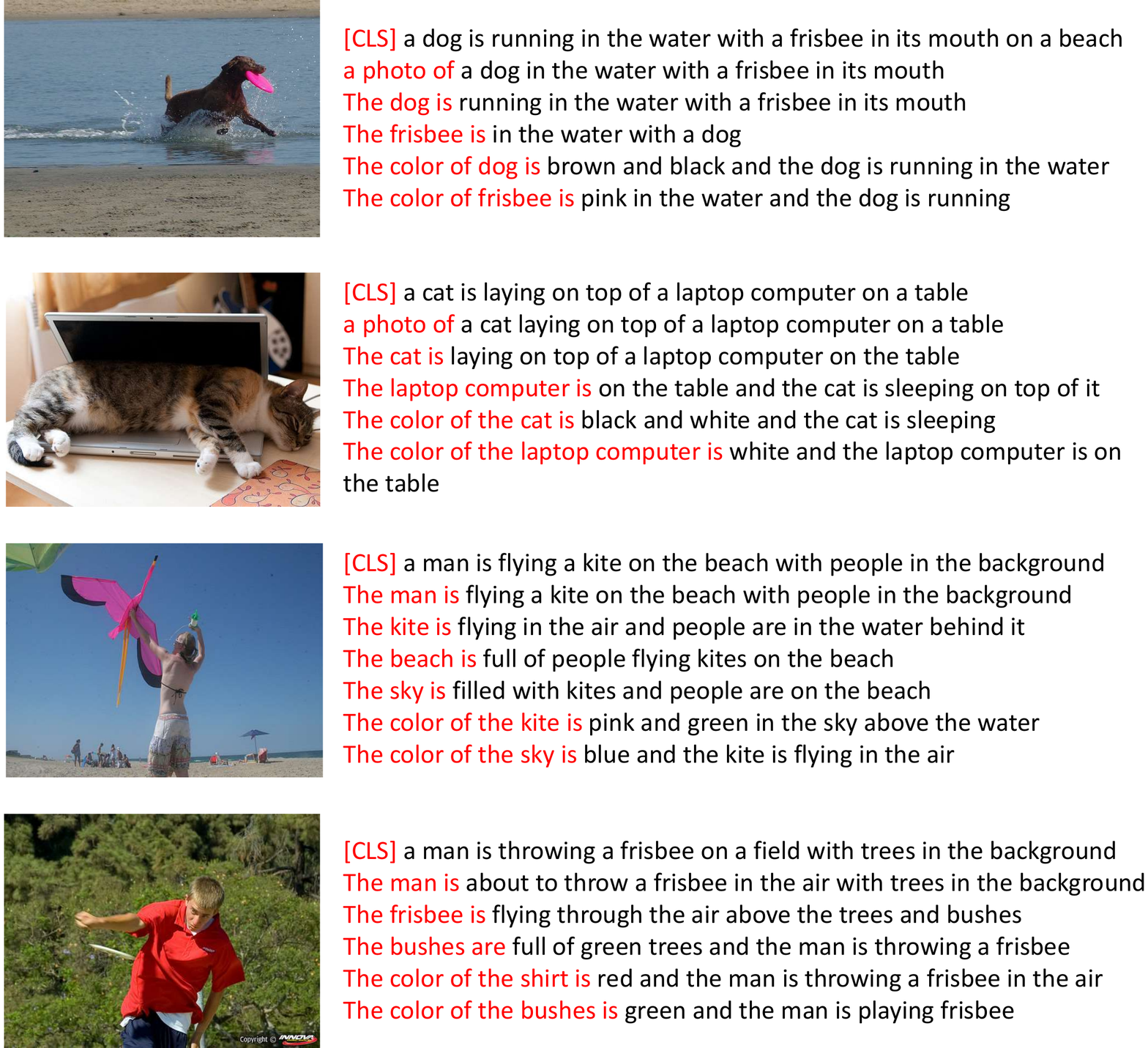}
\end{center}
   \caption{ Captions with different prompts generated by the UniVL. The red text is the prompt. }
\label{fig:cap}
\end{figure*}

\begin{figure*}[t]
\begin{center}
   \includegraphics[scale=0.65]{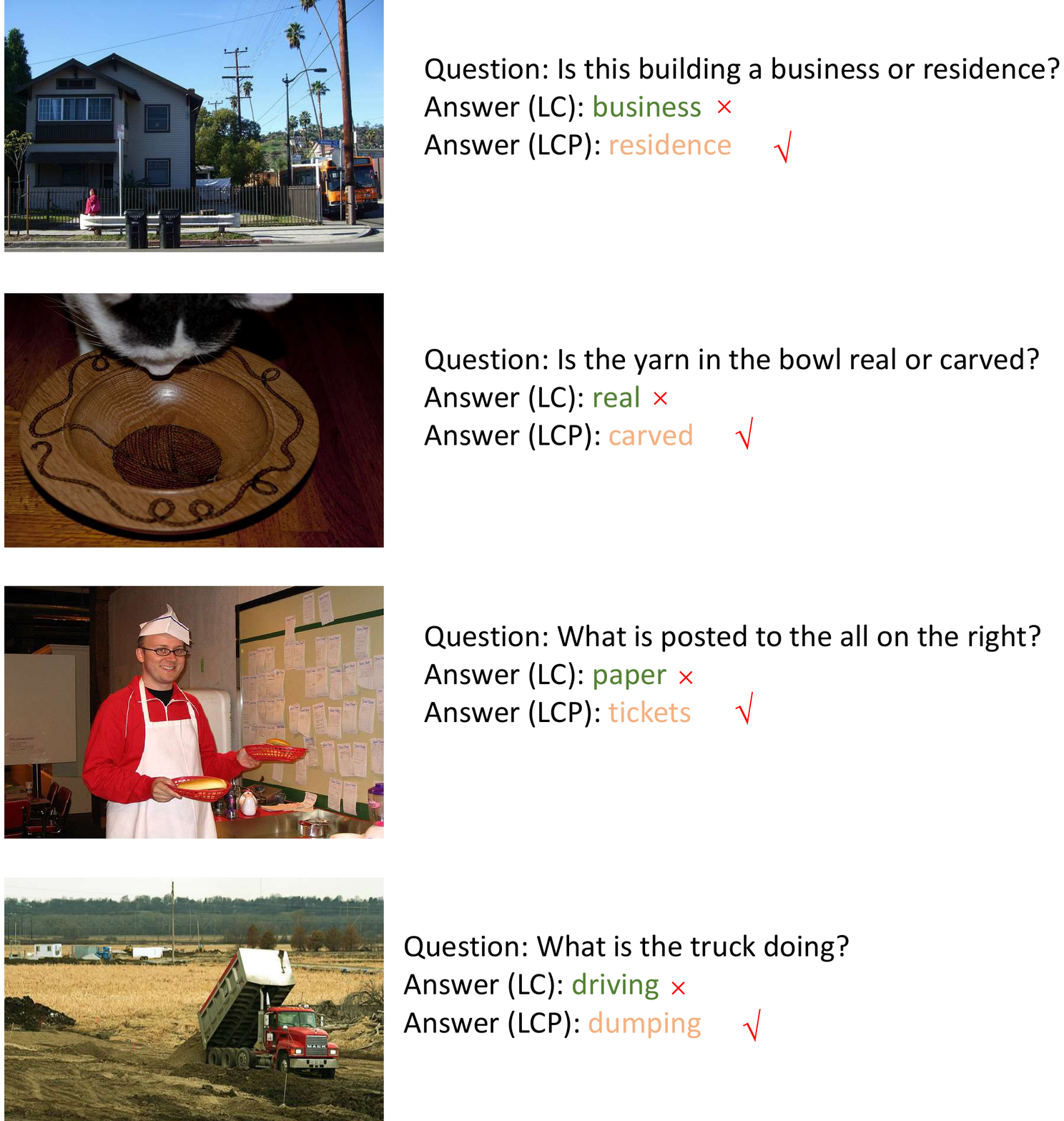}
\end{center}
   \caption{ Results of VQA with questions that have rare answers(not in the predefined answer list). LC, LCP mean linear classifier, learnable context prompt. }
\label{fig:vqa}
\end{figure*}

\begin{figure*}[t]
\begin{center}
   \includegraphics[scale=0.65]{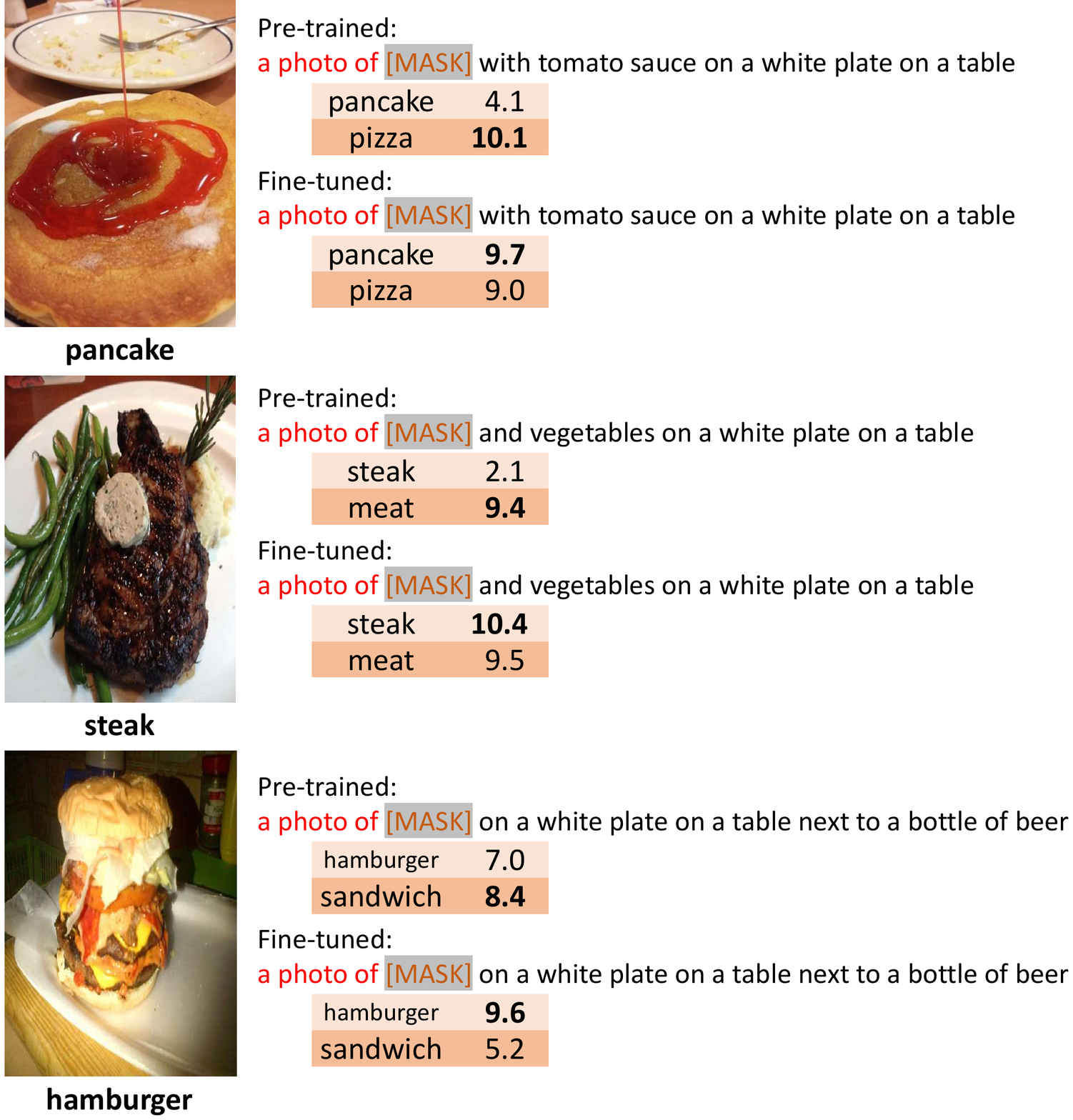}
\end{center}
   \caption{ Results of fine-grained image classification on Food101. }
\label{fig:fic}
\end{figure*}

\subsection{Image Captioning}
We use the special token $\rm [CLS]$ as the prompt in Section5.2 and we show that different natural language prompts can make the model generate descriptions from different perspectives. As shown in Figure \ref{fig:cap}, we feed different prompts to the pre-trained UniVL and since $\rm [CLS]$ is the beginning of a sentence during pre-training, the model can generate a complete caption with $\rm [CLS]$. We change the subject of the sentence, and the generated description changes with the change of the subject. The visualization results show that our pre-trained UniVL learns the correspondence between vision and language and we can control the descriptions generated by the pre-trained model via changing the prompt.

\subsection{Visual Question Answering}
We show that our UniVL can generate open-ended answers and has a significant improvement on questions with rare answers in Section5.3 and Figure \ref{fig:vqa} shows some results. It should be noted that all of the correct answers are included in the expanded answer list, but there are too few training samples with answers that are not in the predefined answer list and it is difficult for the linear classifier to generate correct answers for these rare samples. However, prompt-based method formulates VQA as a text generation task to look more like those solved during pre-training and is a more effective method to unleash the potential of the pre-trained model.

\subsection{Fine-grained Image Classification}
Since we formulate image classification as a text generation task, the model can learn the correspondence between image and the class name, which is more general than one-hot labels. As shown in Figure \ref{fig:fic}, we mask out the food name, and ask the pre-trained model and fine-tuned model to predict the $\rm [MASK]$ token, the pre-trained model without fine-tuned on Food101 fails to generate the correct food class and the fine-tuned model can generate the correct class.

\end{document}

%% file: large_tbl.tex
\begin{table*}[t]
\centering
\begin{tabular}{cccccccccccc}
\toprule
\multirow{3}{*}{\begin{tabular}[c]{@{}c@{}}Number of\\ image-text\\ pairs\end{tabular}} & \multirow{3}{*}{\begin{tabular}[c]{@{}c@{}}$\rm P_{causal}$\end{tabular}} && \multicolumn{4}{c}{Generation}                                                          && \multicolumn{4}{c}{Understanding}                                                             \\ \cmidrule{4-7}\cmidrule{9-12} 
                                                                                        &                                                                        & \multicolumn{4}{c}{\qquad\qquad\quad Image Captioning}                                                    &&& \multicolumn{1}{c}{VE}  & \multicolumn{1}{c}{ITR(R@1)} & \multicolumn{1}{c}{TIR(R@1)} & CLS \\ \cmidrule{4-7}\cmidrule{9-12} 
                                                                                        &                                                                        && \multicolumn{1}{c}{B1} & \multicolumn{1}{c}{B4}       & \multicolumn{1}{c}{R} & C     && \multicolumn{1}{c}{Acc} & \multicolumn{1}{c}{Acc}      & \multicolumn{1}{c}{Acc}      & Acc \\ \cmidrule{1-2}\cmidrule{4-7}\cmidrule{9-12}
\multirow{4}{*}{0.75M}                                                                  & 0.0&                                                                    & \multicolumn{1}{c}{15.7}   & \multicolumn{1}{c}{3.5}      & \multicolumn{1}{c}{10.1}  & 11.7  && \multicolumn{1}{c}{50.9}    & \multicolumn{1}{c}{56.4}         & \multicolumn{1}{c}{43.3}         & 64.3    \\ 
                                                                                        & 0.33  &                                                                 & \multicolumn{1}{c}{50.7}   & \multicolumn{1}{c}{20.2}     & \multicolumn{1}{c}{35.8}  & 68.5&  & \multicolumn{1}{c}{49.6}    & \multicolumn{1}{c}{52.1}         & \multicolumn{1}{c}{41.0}         & 59.7     \\ 
                                                                                        & 0.66&                                                                   & \multicolumn{1}{c}{58.7}   & \multicolumn{1}{c}{23.4}     & \multicolumn{1}{c}{37.5}  & 78.4&  & \multicolumn{1}{c}{47.5}    & \multicolumn{1}{c}{51.8}         & \multicolumn{1}{c}{39.8}         &66.9     \\ 
                                                                                        & 1.0 &                                                                   & \multicolumn{1}{c}{66.9}   & \multicolumn{1}{c}{24.8}     & \multicolumn{1}{c}{38.7}  & 84.9&  & \multicolumn{1}{c}{33.3}    & \multicolumn{1}{c}{41.9}         & \multicolumn{1}{c}{27.7}         & 47.1     \\ \cmidrule{1-2}\cmidrule{4-7}\cmidrule{9-12}
\multirow{4}{*}{1.5M}                                                                   & 0 &                                                                     & \multicolumn{1}{c}{24.9}   & \multicolumn{1}{c}{5.3} & \multicolumn{1}{c}{16.8}  & 18.2 & & \multicolumn{1}{c}{73.5}    & \multicolumn{1}{c}{82.6}         & \multicolumn{1}{c}{70.4}         & 85.4     \\ 
                                                                                        & 0.33  &                                                                 & \multicolumn{1}{c}{59.8}   & \multicolumn{1}{c}{22.9}     & \multicolumn{1}{c}{38.0}  & 73.7&  & \multicolumn{1}{c}{72.4}    & \multicolumn{1}{c}{79.4}         & \multicolumn{1}{c}{67.9}         & 79.9     \\ 
                                                                                        & 0.66  &                                                                 & \multicolumn{1}{c}{65.1}   & \multicolumn{1}{c}{24.8}     & \multicolumn{1}{c}{38.9}  & 82.4 & & \multicolumn{1}{c}{72.2}    & \multicolumn{1}{c}{80.8}         & \multicolumn{1}{c}{69.4}         & 85.1     \\ 
                                                                                        & 1    &                                                                  & \multicolumn{1}{c}{69.4}   & \multicolumn{1}{c}{26.0}     & \multicolumn{1}{c}{39.4}  & 89.7&  & \multicolumn{1}{c}{61.9}    & \multicolumn{1}{c}{70.2}         & \multicolumn{1}{c}{61.3}         &61.7     \\ \cmidrule{1-2}\cmidrule{4-7}\cmidrule{9-12}
3.4M                                                                                    & 0.5 &                                                                   & \multicolumn{1}{c}{\textbf{96.1}}   & \multicolumn{1}{c}{\textbf{35.6}}     & \multicolumn{1}{c}{\textbf{67.0}}  & \textbf{116.8}& & \multicolumn{1}{c}{\textbf{78.1}}    & \multicolumn{1}{c}{\textbf{94.3}}         & \multicolumn{1}{c}{\textbf{82.8}}         & \textbf{92.8}    \\ \bottomrule
\end{tabular}
\vspace{-0.1in}
\caption{Ablation study on different proportions of causal mask and the number of pre-training image-text pairs. $\rm P_{causal}=\frac{causal\ mask}{causal\ mask + bidirectional\ mask}$}
\label{tab:p}
\vspace{-0.1in}
\end{table*}